\documentclass[sigconf, authorversion]{acmart}
\usepackage{booktabs} % For formal tables

\setcopyright{rightsretained}
\acmDOI{10.475/123_4}

\acmISBN{123-4567-24-567/08/06}

\acmConference[WOODSTOCK'97]{ACM Woodstock conference}{July 1997}{El
  Paso, Texas USA}
\acmYear{1997}
\copyrightyear{2016}

\acmArticle{4}
\acmPrice{15.00}

\editor{Jennifer B. Sartor}
\editor{Theo D'Hondt}
\editor{Wolfgang De Meuter}

\begin{document}
\title{Knowledge-Base Enriched Word Embeddings for Biomedical Domain}
% \titlenote{Produces the permission block, and
% \subtitle{Extended Abstract}
% \subtitlenote{The full version of the author's guide is available as
%   \texttt{acmart.pdf} document}

\author{Kishlay Jha}
% \authornote{Dr.~Trovato insisted his name be first.}
\affiliation{%
  \institution{University of Virginia}
  \city{Charlottesville}
  \state{VA}
  \postcode{22903}
}
\email{{kj6ww}@virginia.edu}

\maketitle

\section{Abstract}
\label{sec:abstract}
Word embeddings have been shown adept at capturing the semantic and syntactic regularities of the natural language text, as a result of which these representations have found their utility in a wide variety of downstream content analysis tasks. Commonly, these word embedding techniques derive the distributed representation of words based on the local context information. However, such approaches ignore the rich amount of explicit information present in knowledge-bases. This is problematic, as it might lead to poor representation for words with insufficient local context such as \textit{domain specific} words. Furthermore, the problem becomes pronounced in domain such as bio-medicine where the presence of these domain specific words are relatively high. Towards this end, in this project, we propose a new word embedding based model for biomedical domain that jointly leverages the information from available corpora and domain knowledge in order to generate knowledge-base powered embeddings. Unlike existing approaches, the proposed methodology is simple but adept at capturing the precise knowledge available in domain resources in an accurate way. Experimental results on biomedical concept similarity and relatedness task validates the effectiveness of the proposed approach. 

%given a set of biomedical concepts and a set of articles they occurred

% Community detection techniques to study the bibliographical knowledge bases have gained significant attention recently. While significant advances have been made by the prior studies; yet, a majority of them have essentially focused on static graphs, thereby, neglecting the temporal characteristics of the networks. This is problematic because it is known that the networks in general are dynamic in nature. Furthermore, their effect is especially prevalent in domains such as biomedicine, where some new facts emerge and some are rendered obsolete every now and then. To tackle this challenge, in this preliminary study, we attempt to investigate the evolution of network communities in PubMed (a popular bibliographical database) over the period of time. More specifically, we first split the corpus into distance time scopes, extract the relevant entities and represent them in a bipartite network. We then apply the bipartite stochastic block model~\cite{larremore2014efficiently} to find the hidden communities. Apart from providing a statistically principled solution, they are also robust to possible information loss. In the experiments, we demonstrate the model's ability to identify meaningful community structure for chosen test-cases from the biomedical domain. 

\section{Introduction}
Understanding the semantics behind a text is a fundamental problem in the field of Natural Language Processing (NLP)~\cite{bengio2003neural,bengio2006neural,collobert2011natural}. Towards this end, recently, modelling the distributed representation of words has attracted considerable attention from the research community. This is primarily due to the development of neural network inspired word embedding models~\cite{bengio2003neural,mikolov2013distributed,jha2016mining,jha2018concepts,jha2019hypothesis,jha2016miningA} that have shown to encode the semantic structure of words at a granular level. Simply put, these embedding models learn continuous low dimensional dense vectors of words, commonly known as word embeddings, in a completely unsupervised manner. 
% Commonly, distributed representation of words are learned based on the assumption of Bag of words (BOW)~\cite{zhang2006distributed}, where each individual word is represented as a one-hot vector (i.e., one component in the vector has a value one and rest are zero). However, this representation fails to capture the rich internal semantic structure among words. More recently, studies like ~\cite{bengio2003neural,mikolov2013distributed} have focused on representing words as continuous low dimensional dense vectors. Vectors of this kind, commonly known as word embeddings, have been shown to capture the implicit semantics of the corresponding words at a granular level.
While considerable success has been achieved by these embedding models, yet, they are still afflicted with certain drawbacks. The major being their sole reliance on the assumption of \textit{distributional hypotheses} (words appearing in similar context have similar meaning)~~\cite{harris1954distributional}. This exclusive reliance on the local-context leads to poor representation of words that have insufficient statistical information such as \textit{domain-specific} or \textit{rare} words. Such domain-specific words are present in sizable amount in almost all the domains and particularly in domains such as bio-medicine. As an example, consider a pair of medical concepts "Heart" and "Aorta". Despite the fact that this pair rarely co-occur together in the corpus (less local-context information), they are still semantically related (aorta is the largest artery in the body that is present in heart) in the knowledge-base. This insight leads us to the crux of the problem that we intend to address in this project - How to supplement the data-driven embedding model with the curated information from the knowledge-base so that it supplements the insufficient context information and generates knowledge-base enriched vector representations?

To achieve our goal, we leverage both the corpus based information and the KB available in the biomedical domain. As means of corpus information, we choose MEDLINE \footnote{https://www.nlm.nih.gov/pubs/factsheets/medline.html}, the popular and perhaps the most comprehensive literature repository in the biomedical domain.
% Towards this direction, we leverage MEDLINE\footnote{https://www.nlm.nih.gov/pubs/factsheets/medline.html}, the popular and perhaps the most comprehensive literature repository in the biomedical domain. 
Every article in this repository is indexed with a set of medical concepts. Being manually curated by subject-matter-experts, they are highly precise and accurate. In this project, we use these medical concepts as a unit of representation for documents. The same concepts are also present in the biomedical KB (arranged in the form of a hierarchy) thus providing us with another complementary source of information. 
% structural information about these medical concepts. 
% As our goal in the project is to generate embeddings for these medical concepts, we leverage both the corpus information (MEDLINE) and the domain knowledge (MeSH tree). 
While MEDLINE provides us with the local-context information for the concepts, the hierarchical KB provides complementary information for those words with insufficient local context-information. We would like to note that, unlike the hierarchies present in general domain such as WordNET~\cite{miller1995wordnet}, the chosen KB in biomedical domain is unique in itself with its own peculiarities. The two main distinctive features are: a) the concepts are strictly arranged in an \textit{IS-A} relationship, b) the lower the concept pairs in the taxonomy the more specific information they possess.
% has its own distinct features such as
% it is worthwhile to point that the KB of biomedical domain is unique in itself with its own peculiarities. Unlike the hierarchies present in general domain such as WordNET~\cite{miller1995wordnet}, KB in biomedical domain has its own distinct features such as: a) one concept in biomedical KB may belong to several sub-tress, b) the lower the concept pairs in the taxonomy the more shared information they posses. 
This uniqueness causes the straight-forward portability of existing works~\cite{muneeb2015evaluating,chiu2016train,yu2014improving} to have a limited effect. More specifically, most of the prior works have attempted to augment embeddings with relational/categorical knowledge. The structured representation of these knowledge bases, commonly in the form of triplets (subject-predicate-object), are distinct than the current resource of interest - hierarchical KB. Thus, it is unclear on how the prior works could be applied to the biomedical domain in their plain vanilla form. Motivated with this narrative, in this project, we explore the distinctive features of biomedical KB to generate embeddings that are more suited and optimized for biomedical specific applications. Our main technical challenge lies in jointly modelling the local context information from the natural language text and structural information from human curated KB. To accomplish this objective, we propose a deep-learning based solution that simultaneously exploits the two complementary sources of information. In particular, the proposed model is essentially a recurrent neural networks based concept language model trained on sequential text taken from both unstructured corpora (local context) and structured taxonomy (domain knowledge). To validate the effectiveness of proposed model, we performed experiments on biomedical concept similarity/relatedness task.

To summarize, in this study, we made the following particular contributions:
%\yell{Needs review}
\begin{enumerate}
      \item We proposed a new word embedding model for biomedical domain that jointly exploits both the corpus based information and hierarchical knowledge-base to produce knowledge-base augmented word embeddings.
      
      \item Compared to the existing approaches, the proposed model is relatively simple but adept at capturing the explicit semantic knowledge in an accurate fashion.
\end{enumerate} 
%We use RNNs, in a novel way to model both the local context and the hierarchical knowledge.

\section{Related Work}

\label{sec:relatedwork}
Improving the distributed representation of words is an important problem in the research area of natural language processing~\cite{collobert2008unified, bengio2003neural, bengio2006neural}. For a recent survey, please refer~\cite{li2018word, almeida2019word}. In our chosen domain of interest (i.e., biomedical domain), recent years have witnessed some early attempts towards applying word embedding models for several bioNLP tasks. Chiu et al.~\cite{chiu2016train} studied and reported the effects of input corpora, dimension size, parameters on the quality of embeddings. Similarly, \cite{muneeb2015evaluating} conducted intrinsic evaluations on multitude of linguistic tasks such as Part-of-Speech tagging, chunking, named entity recognition, mention detection.

% Munnet et al. \cite{muneeb2015evaluating} trained both the Skip-gram and CBOW models over the PubMed Central Open Access (PMC) corpus with approximately 400 million tokens. On the task of semantic similarity and relatedness, they report that Skip-gram model (word2vec) performed the best for the task of semantic similarity, on the other hand, none of the models outperformed others in the semantic relatedness task. Chiu et al.~\cite{chiu2016train} performed analysis on the effect of input corpora, architecture and hyperparameter setting (negative sample size, sub-sampling, minimum-count, learning rate, vector dimension, and context window size) on the quality of embeddings. 
% In their results, they report the values of some of the influential hyperparamater, 10 (negative sample size), 1e-4 (sub-sampling), 0.05 (learning rate), 200 (vector dimension), and 30 (context window size) for word2vec and conclude that the size of corpora does not affect the quality of word embeddings. Similar to the aforementioned work, a more recent study \cite{zhu2017semantic} examined the effect of recency, size and section of biomedical publication (abstract/full-text) data on the performance of word2vec. They reported that the models trained on recent datasets did not boost the performance and as compared to the full text articles bodies, abstracts excel in accuracy.

While these studies were important in demonstrating the salient aspects of biomedical embedding models, they did not focus on making model level innovation. Furthermore, the quality of embeddings generated by these models for domain-specific words are relatively poor~\cite{zhu2017semantic}. To overcome this, in this study, we leverage the rich domain knowledge present in biomedical domain and explore effective ways to integrate them. We would like to note that there have been studies conducted in NLP domain that integrate certain forms of prior knowledge~\cite{celikyilmaz2010enriching,wang2017integrating}. In ~\cite{yu2014improving}, the authors proposed a simple but effective method to encode relational knowledge. In particular, they extended the objective function of skip-gram by incorporating the prior knowldege in the form of a regularizer. Likewise, in another related study~\cite{liu2015learning}, the authors proposed an alternate method to encode the semantic knowledge via ordinal constraints. Although these models made important advances, their application is limited to the biomedical domain. This is mainly due to the unique challenges posed by the hierarhical KB present in biomedical domain. 

\section{Model Architecture} 

In this section we discuss our proposed model to generate word embedding for biomedical domain. A complete pipeline of the proposed framework is illustrated in Figure \ref{fig:proposedModel}.  

\subsection{Continuous Bag-of-Words Model}
In general, the word embeddings are generated using neural networks with majority of them modelling the objective function as a one trying to predict either the word under consideration based on a context described through a window or the vice-versa. The two most popular word embedding models are Continuous Bag-of-Words Model (CBOW) and Skip-gram models~\cite{mikolov2013distributed,mikolov2013efficient}. Both CBOW and Skip-gram models are three-layer neural networks, containing input, projection, and output layers. In this project, we base our proposed model on CBOW. The CBOW model learns word embedding by using context words to predict the center word $w_{t}$, where the context words refer to the neighboring words within a window size $c$ near the center word in a sentence. Given a sequence of training words $w_{1}$, $w_{2}$,..., $w_{T}$, the CBOW model has the following objective function:

\begin{equation}
J_{1} = \frac{1}{T}{\sum_{t=1}^T} \log p(w_{t}|w_{t-c},...,w_{t-1}, w_{t+1},...,w_{t+c}).
\end{equation}

The CBOW model first computes the hidden layer $h_t$ for word $w_t$ by averaging the input embeddings for its context words.

\begin{equation}
h_t =  \frac{1}{2c}{\sum_{-c<j<c,j\neq 0}} v_{t+j},
\end{equation}
where $v_{i}$ is the input representation for word $w_{i}$. Then $p(w_{o}|w_{t-c},...,w_{t-1}, w_{t+1},...,w_{t+c})$ is calculated based on hidden state $h_t$ using a softmax function:

\begin{equation}
    P(w_o|h_t) = \frac{\exp (v_o^{'T}h_t)}{\sum_{i=1}^N \exp(v_i^{'T}h_t) },
\end{equation}
where $v_i^{'}$ is the output representation for word $w_{i}$. Note that, the representation vector for $v_n$ is between input layer and projection layer. $v_n'$ denotes representation vector between projection layer and output layer.

% The probability $p(w_{t}|w_{t+j})$ is calculated using a softmax function:

% $$ P(w_t | w_{t+j}) = \frac{exp(v_t^{'T}v_{t+j})}{\sum_{t=1}^N exp(v_t^{'T}v_{t+j}) }$$ where $v_n$ and $v_n'$ are input and the output representation vectors of word $w_n$, and N in the total vocabulary size.  Note that, the representation vectors $v_n$ are between input layer and projection layer, and $v_n'$ are between projection layer and output layer. 

\begin{figure}
\centering
\includegraphics[width=3.5in]{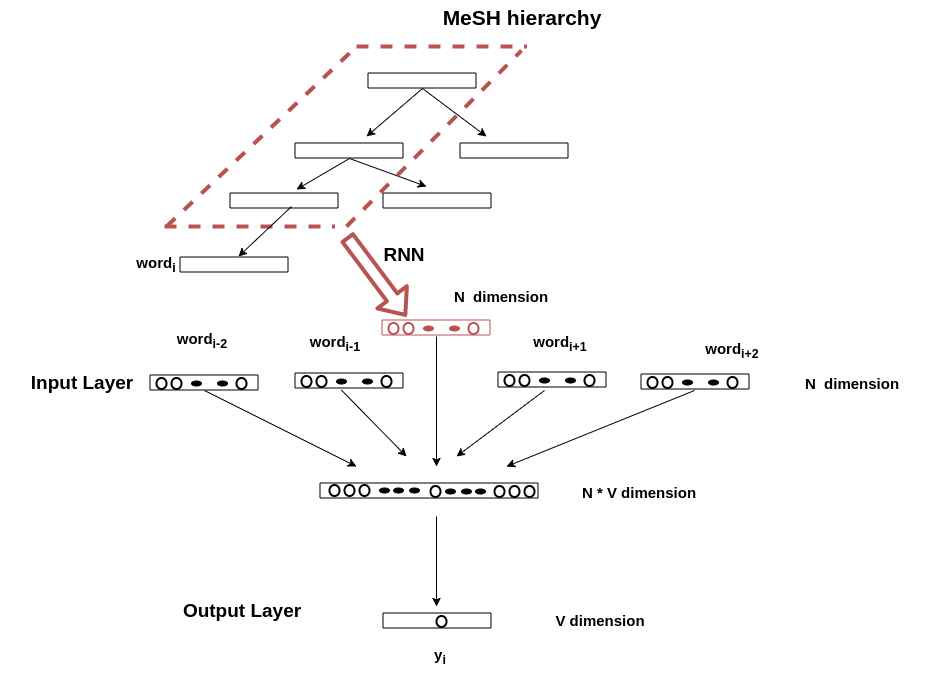}
\caption{Basic architecture of the proposed model}
\label{fig:proposedModel}
\end{figure}

In this project, we use CBOW to incorporate the corpus based information into the model. Having incorporated the local context, the next step is to integrate the domain knowledge from external KB. As mentioned before, the chosen KB is hierarchical in nature. Consequently, every medical concept is associated with a sequence of ancestors. To model this sequential nature of ancestors, we use recurrent neural network (RNN).
% Naturally, this leads to the problem presenting itself as a sequence modelling task, we use recurrent neural networks to model them. 
The hidden layer of RNN contains the history of ancestors linked as an input to the word-level recurrent neural network to predict the next word together with the word-level history vector. This allows the language model to predict the next word probability distribution beyond the words in the current sentence. In Figure \ref{fig:proposedModel}, RNN could be any RNN variants, such as Long Short Term Memory (LSTM)\cite{hochreiter1997long} and Gated Recurrent Unit (GRU). In our implementation, we use LSTM to deal with the vanishing gradient problem.

\subsection{Long Short-Term Memory}
LSTMs address the vanishing gradient problem commonly found in RNNs by incorporating gating functions into their state dynamics. At each time step, an LSTM maintains a hidden vector $h$ and a memory vector $m$  responsible for controlling state updates and outputs. More concretely, we define the computation at time step $t$ as follows:

\begin{equation}
\begin{split}
	g^u = \sigma( W^u *  h_{t-1} +  I^u * x_t) \\
    g^f = \sigma( W^f *  h_{t-1} +  I^f * x_t) \\
	g^o = \sigma( W^o *  h_{t-1} +  I^o * x_t) \\
	g^c = \tanh( W^c *  h_{t-1} +  I^c * x_t) \\
	m_t =  g^f \odot  +   g^u \odot  g^c \\
	h_t = \tanh( g^o \odot  m_{t-1}) 
\end{split}
\end{equation}
Here $\sigma$ is the logistic sigmoid function, $ W^u,  W^f,  W^o,  W^c$ are recurrent weight matrices, and $ I^u,  I^f,  I^o,  I^c$ are projection matrices.

\subsection{Proposed Model}
Our proposed model incorporates the external knowledge-base as external information to achieve more accurate embeddings. The architecture of our proposed model is shown in Figure \ref{fig:proposedModel}. For every medical concept (or term), we treat its ancestor nodes obtained from the  hierarchy as a sequence data, and then apply LSTM model on this sequence. The outputs from the last layer are provided as a hierarchy context information for target medical concept. The local context words in corpus and hierarchy context information are concatenated together to be feed into the CBOW model to predict the corresponding target medical concept. By this mechanism, our proposed approach simultaneously leverages both local context information and explicit knowledge from KB in an effective manner.

% \begin{figure}
% \centering
% % \includegraphics[width=4in]{proposal/CBOW.png}
% \includegraphics[height = 6cm,width=6cm]{proposal/CBOW.png}
% \caption{The CBOW architecture (Adapted from \cite{mikolov2013distributed,mikolov2013efficient})}
% \label{fig:cbow}
% \end{figure}

\section{Experiments}

\label{sec:experiments}

Having explained the methodological details, we now empirically evaluate, analyze and discuss the proposed model's performance on biomedical concept similarity/relatedness task. To obtain medical concept embeddings, we set its embedding dimension as 100, learning rate $\epsilon=0.001$, batch size as 100 instances, and the model is trained for 100 epochs to report the results.

\subsection{Dataset Preparation}

To examine the applicability of the model, we sampled a subset of entire dataset (MEDLINE). For this project, the total number of article used from MEDLINE are 100,000. The proposed model was trained on this subset. Before we delve into the details of our evaluation, we briefly describe our unit of representation (MeSH terms) and chosen external knowledge base.

\subsection{Medical Subject headings (MeSH)}

Medical Subject Headings (MeSH) are National Library of Medicine (NLM) controlled vocabulary that human experts use to index journal articles in the life sciences domains. Mesh terms are classified into three categories a) Descriptors, b) Qualifiers and c) Supplementary concept records. \textit{Descriptors} represent the conceptual meaning of the article. In this work, we use Descriptors as the unit of representation for documents.

\subsection{External knowledge base }
Our chosen KB is MeSH hierarchy. MeSH terms are organized in the form of a tree hierarchy based on their level of specificity. Every MeSH term has a corresponding tree code which represents its level of specificity in the tree. As an example, the tree code for concept \textit{Migraine Disorders} is: C10.228.140.546.399.750. 

\subsection{Evaluation Datasets}

To evaluate the output embeddings on biomedical concept similarity/relatedness task, we borrow evaluation set from~\cite{aouicha2016computing}. The description of both datasets are provided below:

% Table~\ref{benchmarkSet} enumerates the benchmark datasets along with the number of concept pairs that were manually rated by human experts to denote semantic similarity. 

\begin{itemize}
    \item \underline{MeSH-1} : The first dataset (MeSH-1) \cite{pedersen2007measures} was created by experts from Mayo Clinic and consists of a set of word pairs that are related to general medical disorders. The similarity of each concept pair was assessed by 3 physicians and 9 medical coders. Each pair was annotated on a 4 point scale: practically synonymous, related, marginally, and unrelated. The average correlation between physicians is 0.68 and between experts is 0.78. In this project, we use 19 concepts that were present in our vocabulary as our evaluation set.  
    \item \underline{MeSH-2}: The second biomedical benchmark (MeSH-2) was introduced in ~\cite{hliaoutakis2005semantic}. It consists of a set of 36 word pairs extracted from the MeSH repository. The similarity between word pairs was assessed by 8 medical experts and assigned a score between 0 (non-similar) to 1 (synonyms). In this project, we use 20 concepts that were present in our vocabulary as our evaluation set.
\end{itemize}

% \subsection{Evaluation metric}
% The common way of examining the quality of word embeddings is to analyze their correlation coefficient with human experts. The higher the correlation, the better it is assumed to captures the human intuition of natural language. Consequently, we use Spearman coefficient as the evaluation metric.
 
%  \begin{equation}
% % \begin{split}
% \rho = 1 - \frac{6 \sum d_{i}^2 }{n({n}^2-1)}
% % \end{split}
% \label{eq:spearman}
% \end{equation}

% \underline{Spearman coefficient} (\(\rho\)):  This metric is used to correlate word pair rankings produced by the proposed method to the ones assigned by expert judgments. The formula for calculating \(\rho\) is given in Equation \ref{eq:spearman}, where \({d_i}\) is the difference between the ranks of \({x_i}\) and \({y_i}\), \({x_i}\) refers to the \(i^{th}\) element in the list of human judgments, \({y_i}\) to the corresponding \(i^{th}\) element in the list of semantic similarity computed values, and \textit{n} is the total number of word pairs. In this work, we use Spearman coefficient to judge the quality of our result.

\subsection{Results}

Table \ref{table:mesh1} and Table \ref{table:mesh2} reports the Spearman (\( \rho \)) coefficient values obtained by applying the proposed model on the datasets MeSH-1 and MeSH-2 respectively. In order to have a baseline for comparison, we generated word embeddings using only the domain knowledge and corpus information respectively. As it can be observed from the table \ref{table:mesh1} and table \ref{table:mesh2}, the proposed model outperforms the baseline and achieves a higher correlation with human experts.

% \begin{table}[!t]
% \centering
% \caption{Summary of datasets used for evaluating semantic similarity/relatedness task for biomedical concept pairs.}
% \label{benchmarkSet}
% \begin{tabular}{|p{2.8cm} |c|c} \hline
% Datasets & Concept Pairs\\  \hline
% MeSH-1\cite{pedersen2007measures} & 19 \\ \hline
% MeSH-2\cite{hliaoutakis2005semantic} & 20 \\ \hline
% \end{tabular}
% \end{table}

\underline{Discussion:} Although the proposed model outperforms the baseline the overall Spearman coefficient is relatively low. We believe the reason for this lies in the insufficient training. As mentioned before, due to limited time constraint and computational challenges, the training was performed on a small subset of entire dataset. In future, we intend to train of model on the entire massive dataset. One insight to note from the experiments is the higher correlation of proposed model. This result validates the importance of incorporating both local context and explicit semantic knowledge to generate semantically meaningful embeddings. Furthermore, the higher correlation also points the effectiveness of model in capturing the human intuition of similarity/relatedness.

\begin{table}[!t]
\centering
\caption{Correlation values relative to human judgments for- MeSH-1}
\begin{tabular}{| p{5.00cm} |c|c|} \hline
Methods & Spearman coefficient\\  \hline
Domain knowledge only & 0.228 \\ \hline
CBOW & 0.304 \\ \hline
\textbf{Proposed Model} & \textbf{0.312} \\ \hline
\end{tabular}
\label{table:mesh1}
\end{table}

\begin{table}[!t]
\centering
\caption{Correlation values relative to human judgments for- MeSH-2}
\begin{tabular}{| p{5.00cm} |c|c|} \hline
Methods & Spearman coefficient\\  \hline
Domain knowledge only & 0.280 \\ \hline
CBOW & 0.363 \\ \hline
\textbf{Proposed Model} & \textbf{0.371} \\ \hline
\end{tabular}
\label{table:mesh2}
\end{table}

\section{Conclusion}

In this project, we proposed a new recurrent neural network based language model for biomedical domain that leverages both corpus based contextual information and explicit semantic knowledge (present in the form of hierarchy) to produce high-quality word embeddings. The fusion of taxonomical knowledge  that are hand-engineered by subject matter experts into corpus based embedding model provides additional semantic evidence to words with poor local context, thereby, enriching the quality of overall embeddings. Experiments on biomedical concept similarity/relatedness task have illustrated that the proposed knowledge-based powered approach significantly improves the quality of word representation. In future work, we intend to explore more effective ways of integrating this knowledge base and to develop strategies in order to make the system computationally efficient.

% \begin{table*}
% \centering
% \caption{Tentative execution plan}
% \label{my-label}
% \begin{tabular}{|l|l|} \hline
% \textbf{Month} & \textbf{Milestone} \\ \hline
% Feburary & Define the problem formally and study literature\\ \hline
% March & \begin{tabular}[c]{@{}l@{}}Create network and analyze small world properties such as  (clustering coefficient, modularity)\end{tabular} \\ \hline
% April & Propose and implement model \\ \hline
% May & Report writing \\
% \hline \end{tabular}
% \end{table*}

% \section{Tentative Execution Plan}
% \label{sec:executionPlan}

%\end{document}  % This is where a 'short' article might terminate

\bibliographystyle{ACM-Reference-Format}
\bibliography{sample-bibliography}

\end{document}